# An LQR-assisted Control Algorithm for an Under-actuated In-pipe Robot in Water Distribution Systems

Saber Kazeminasab
Department of Electrical and Computer Engineering- Texas A&M University
College Station, Texas, USA
skazeminasab@tamu.edu

Roozbeh Jafari
Departments of Electrical and Computer, Computer Science, and Biomedical Engineering - Texas A&M University
College Station, Texas, USA
rjafari@tamu.edu

M. Katherine Banks
College of Engineering- Texas A&M University
College Station, Texas, USA
k-banks@tamu.edu

## ABSTRACT

**To address the operational challenges of in-pipe robots in large pipes of water distribution systems (WDS), in this research, a control algorithm is proposed for our previously designed robot [4]. Our size adaptable robot has an under-actuated modular design that can be used for both leak detection and quality monitoring. First, nonlinear dynamical governing equations of the robot are derived with the definition of two perpendicular planes and two sets of states are defined for the robot for stabilization and mobilization. For stabilization, we calculated the auxiliary system matrices and designed a stabilizer controller based on the linear quadratic regulator (LQR) controller, and combined it with a proportional-integral-derivative (PID) based controller for mobilization. The controller scheme is validated with simulation in MATLAB in various operation conditions in three iterations. The simulation results show that the controller can stabilize the robot inside the pipe by converging the stabilizing states to zero and keeping them in zero with initial values between -25° and +25° and tracking velocities of 10cm/s, 30cm/s, and 50cm/s which makes the robot agile and dexterous for operation in pipelines.**

## 1 INTRODUCTION

Water Distribution System (WDS) is responsible to deliver potable water to residential areas. An intentional or accidental incident to pipelines causes damage which results in leak and water loss in the network [1, 5]. So, it is required to assess the condition of pipes periodically and in case of a leak, localize the leak [2]. Also, the quality of potable water in WDS needs to be monitored to ensure the "health" of water [6]. In-pipe robots are promising solutions to this aim [3]. However, there are a lot of uncertainties and disturbances in pipelines since a pressurized flow with high speed is present and the pipe geometry is not accurately known [7]. These uncertainties and disturbances make the operation of in-pipe robots in large pipelines an extremely challenging task.

In this paper, we present concept of a motion controller algorithm that stabilizes the robot during operation while the robot tracks a desired velocity. The paper is organized as follows. In section 3, a brief explanation of the robot and modeling are presented. In section 4, the stabilizing and velocity tracking controllers are developed. In section 5, the controller is simulated in MATLAB Simulink environment, and in the paper is concluded in section 6.

## 3 ROBOT DESIGN

The robot is designed to have a streamlined shape. Three adjustable arm modules that are equipped with actuator modules are anchored on a central processor with a 120° angle. The robot's flexible mechanism enables it to have variable size to adapt to the pipe diameter. Passive springs facilitate the adjustability of the arms in which they connect to the arms and the central processor with pins. Actuator modules locate at the end of the arms and provide traction force for the robot. Actuator modules are gear-motors and equipped with encoders to measure the motors' shaft velocity. The actuators are located and sealed on the arms. The central processor is composed of two hemispheres that are fastened and sealed together with an O-ring. The electronic components are located inside the central processor and the battery is located below the central processor and provides power for the robot's components. In Fig. 1, the overall view of the robot and its components are shown. More details on the robot, characterization, and fabrication can be found in [4]. The robot and modeling parameters are described in Table 1. These parameters are shown in Fig. 1.

### 3.1 Robot Modeling

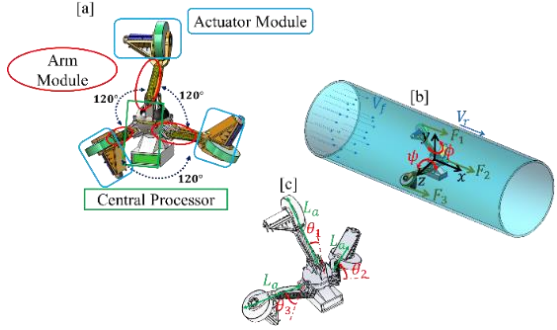

**Figure 1:** Robot design and modeling. a) CAD design and components, b) Free body diagram of the robot in the pipe. c) Geometrical representation of the robot.

**Table 1: Robot specifications.**

| Parameter | Description | Value |
|---|---|---|
| $\theta_1$ | Arm #1 angle | Variable |
| $\theta_2$ | Arm #2 angle | Variable |
| $\theta_3$ | Arm #3 angle | Variable |
| $L_a$ | Arm length | $17 cm$ |
| $m$ | Robot mass | $2.23\ kg$ |
| $I_{yy}$ | Robot's moment of inertia around the y-axis | $0.0126 kg.m^2$ |
| $I_{zz}$ | Robot's moment of inertia around the z-axis | $0.0093 kg.m^2$ |
| $\phi$ | Robot's rotation angle around the y-axis | Variable |
| $\psi$ | Robot's rotation angle around z-axis | Variable |
| $F_1$ | The traction force of actuator #1 | Variable |
| $F_2$ | The traction force of actuator #2 | Variable |
| $F_3$ | The traction force of actuator #3 | Variable |
| $F_d$ | Drag force | Variable |
| $\tau_1$ | Torque for actuator #1 | Variable |
| $\tau_2$ | Torque for actuator #2 | Variable |
| $\tau_3$ | Torque for actuator #3 | Variable |
| $R$ | Wheel radius | $5 cm$ |
| $D$ | Pipe diameter | Variable |
| $V_f$ | Flow velocity | Variable |
| $V_r$ | Robot velocity | Variable |

The dominant force applied on the robot is drag force [8], $F_d$, and is computed as follows:

$$F_d = \tfrac{1}{2}\rho C_d A (V_r - V_f)^2 \quad (1)$$

where $\rho = 1000\frac{kg}{m^3}$, $C_d$, and $A$ are water density, the robot's drag coefficient (computed 0.47 with computational fluid dynamic (CFD) in our previous work), and the frontal area of the robot facing the flow. The linear robot velocity is along the pipe axis which is the x-axis. Hence, $V_r = \dot{x}$. To derive the mathematical representation of the robot's dynamic, we consider two perpendicular planes, $P_1$ and $P_2$, on the robot that shows the robot's size limitation by pipe diameter in Fig. 2. We have:

$$F_1 + F_2 + F_3 + F_d = m\ddot{x} \rightarrow (F_i = \tfrac{\tau_i}{R}) \rightarrow \tfrac{\tau_1}{R} + \tfrac{\tau_2}{R} + \tfrac{\tau_3}{R} + \tfrac{1}{2}\rho C_d A(\dot{x} - V_f)^2 = m\ddot{x} \quad (2)$$

$$\sum M_y = I_{yy}\ddot{\phi} \rightarrow \tfrac{1}{2}F_3 L_a \cos(\theta_3 - \phi) - \tfrac{1}{2}F_2 L_a \cos(\theta_2 + \phi) = I_{yy}\ddot{\phi} \rightarrow \tfrac{1}{2}\tfrac{\tau_3}{R}L_a \cos(\theta_3 - \phi) - \tfrac{1}{2}\tfrac{\tau_2}{R}L_a \cos(\theta_2 + \phi) = I_{yy}\ddot{\phi} \quad (3)$$

$$\sum M_z = I_{zz}\ddot{\psi} \rightarrow \tfrac{\sqrt{3}}{2}F_3 L_a \cos(\theta_3 - \phi)(1 + \sin\psi) + \tfrac{\sqrt{3}}{2}F_2 L_a \cos(\theta_3 + \phi)(1 + \sin\psi) - F_1 L_a \cos(\theta_1 + \psi) = I_{zz}\ddot{\psi} \rightarrow \tfrac{\sqrt{3}}{2}\tfrac{\tau_3}{R}L_a \cos(\theta_3 - \phi)(1 + \sin\psi) + \tfrac{\sqrt{3}}{2}\tfrac{\tau_2}{R}L_a \cos(\theta_3 + \phi)(1 + \sin\psi) - \tfrac{\tau_1}{R}L_a \cos(\theta_1 + \psi) = I_{zz}\ddot{\psi} \quad (4)$$

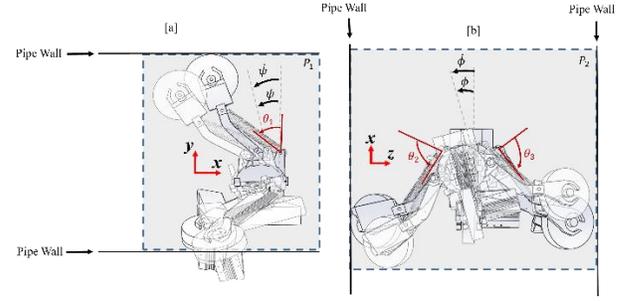

**Figure 2:** Stabilizing states of the robot and the robot size limitation representation with two perpendicular planes. [a] Plane $P_1$, and the relation between $\theta_1$ and $\psi$. [b] Plane $P_2$ and the relation between $\theta_2$, $\theta_3$, and $\phi$.

## 4 STABILIZING AND VELOCITY TRACKING CONTROLLER

In controller design, our goal is to stabilize the robot inside the pipe while it tracks the desired velocity. We define $x_s = [\phi \ \dot{\phi} \ \psi \ \dot{\psi}]^T$ as stabilizing states in Fig. 2, and $x_t = \dot{x}$ for odometry. Following, we design a stabilizer velocity tracker controller based on the mentioned states.

### 4.1 Stabilizer Controller based on the linear quadratic regulator (LQR) Controller

*4.1.1 LQR Concept.* We define some notations in Table 2 that are used in LQR control.

The goal in LQR control is to minimize a cost function:
$$J(K) = \int_0^\infty x^T Q x + u^T R u \ dt \quad (5)$$

To satisfy (5), the control signal is computed as:
$$u = -Kx \quad (6)$$

The gain matrix is computed with:
$$K = R^{-1} B^T P \quad (7)$$

And matrix $P$ is computed with the algebraic Riccati equation:

**Table 2: LQR parameter specifications.**

| Parameter | Description |
|---|---|
| $x$ | System state vector in state-space representation. |
| $u$ | System input vector, $u = [\tau_1 \quad \tau_2 \quad \tau_3]^T$ |
| $Q$ | The nonnegative definite matrix that weights state vector. |
| $R$ | The positive-definite matrix that weights the input vector. |
| $K$ | Gain matrix. |
| $A$ | System matrix. |
| $B$ | Input matrix |
| $P$ | The solution to the algebraic Riccati equation. |

$$-PA - A^T P - Q + PBR^{-1}B^T P = 0 \quad (8)$$

*4.1.2 LQR based Stabilizer Controller.* To design the LQR controller, the nonlinear dynamical equations, (3) and (4) aredefined as $\dot{x}_s = F(x_s, u)$ and linearized around equilibrium point with Taylor series, $x_s^e = [0 \ 0 \ 0 \ 0]^T$ and $u_e = [0 \ 0 \ 0]^T$:

$$\dot{x}_s = F(x_s, u) = F(x_s^e + \delta x_s, u_e + \delta u) \quad (9)$$

We can write:

$$\delta \dot{x}_s = \frac{\partial F}{\partial x_s} \delta x_s + \frac{\partial F}{\partial u} \delta u + O(x_s) + O(u) \quad (10)$$

$O(x_s)$ and $O(u)$ are higher-order terms in Taylor expansion and we omit them. We have:

$$\delta \dot{x}_s = A \delta x_s + B u \quad (11)$$

From (11):

$$A = \frac{\partial F}{\partial x_s}|_{x_s^e} \quad (12)$$

$$B = \frac{\partial F}{\partial u}|_{u_e} \quad (13)$$

The auxiliary system matrices are computed as:

$$A = \begin{bmatrix} 0 & 1 & 0 & 0 \\ 0 & 0 & 0 & 0 \\ 0 & 0 & 0 & 1 \\ 0 & 0 & 0 & 0 \end{bmatrix} \quad (14)$$

$$B = \begin{bmatrix} 0 & 0 & 0 \\ 0 & -\frac{\sqrt{3}D}{4I_{yy}R} & \frac{\sqrt{3}D}{4I_{yy}R} \\ 0 & 0 & 0 \\ -\frac{D}{2I_{zz}R} & \frac{\sqrt{3}D}{4I_{zz}R} & \frac{\sqrt{3}D}{4I_{zz}R} \end{bmatrix} \quad (15)$$

The developed LQR controller stabilizes $x_s$ during operation.

*4.1.3 Velocity Controller.* The wheels of the robot during a straight path have approximately equal angular velocity and the relation between the wheels' angular velocity, $\omega$, and the robot's velocity, $V_r$, is computed as:

$$V_r = R\omega \quad (16)$$

By designing three proportional-integral-derivative (PID) controllers for controlling each wheel's angular velocity, the robot's velocity in a straight path is controlled. The orientation and the angular velocities of the wheels are given to the controller with the observer architecture shown in Fig.3; in which three encoders measure angular velocities of the wheels and the inertial measurement unit (IMU) provide orientation of the robot. The developed controller in 4.1.2 and 4.1.3 are combined and the stabilizer and velocity tracking controller is formulated. In the next section, the performance of the developed controller is evaluated with simulations.

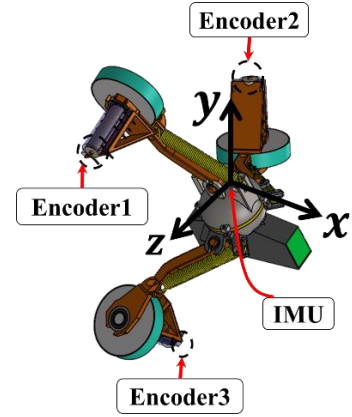

**Figure 3:** Observer architecture for the stabilizer and velocity tracking controller. Three encoders provide odometry for the robot and the IMU at the center of the central processor provides orientation of the robot.

**Table 3: PID controller parameters**

| Parameter | Description | Value |
|---|---|---|
| $K_P$ | Proportional gain | 8.7313 |
| $K_D$ | Integral gain | 0.0072 |
| $K_I$ | Derivative gain | 322.4160 |

## 5 PERFORMANCE OF THE CONTROLLER IN SIMULATION

The proposed control scheme is simulated in MATLAB Simulink toolbox. The system dynamical behavior is presented as nonlinear governing equation, (2)-(4) in a user-defined function. The parameters of the PID controllers are tuned with transfer function method. The tuned parameters are listed in Table 3.

For the LQR controller, $A$, $B$ are from (14) and (15), respectively. $Q = \begin{bmatrix} 200 & 0 & 0 & 0 \\ 0 & 10 & 0 & 0 \\ 0 & 0 & 200 & 0 \\ 0 & 0 & 0 & 10 \end{bmatrix}$ and $R = \begin{bmatrix} 1 & 0 & 0 \\ 0 & 1 & 0 \\ 0 & 0 & 1 \end{bmatrix}$. The gain matrix value is: $K = \begin{bmatrix} 0 & 0 & -11.5470 & -2.5889 \\ -10 & -2.2442 & 5.7735 & 1.2945 \\ 10 & 2.2442 & 5.7735 & 1.2945 \end{bmatrix}$. The

Simulation is repeated for three iterations and the results of each iteration are shown in Fig. 4 in which linear velocity is shown in Fig. 4a, $\phi$ in Fig. 4b, $\psi$ in Fig. 4c, and input control signals for iteration 3 in Fig. 4d. In Table 4, the conditions for each iteration are described. The results show that the developed controller enables the robot to reach the desired linear velocity in less than 0.5 seconds and stabilizes the stabilizing states, Fig. 2, in 1 second. The control input signals, $[\tau_1 \quad \tau_2 \quad \tau_3]^T$, for iteration 3 (where the desired velocity is 0.5m/s and initial values of stabilizing states are larger than iteration 1 and 2) in Fig. 4d, starts from 12000N.mm in $t = 0$ and converges to zero in 0.02 seconds of simulation and fluctuate with 30 N.mm margins. The customized gear motor in this robot can provide this amount of torque with a small drawn electrical current (based on the operating point provided by the manufacture). As the power supply in this robot is a battery, the drawn current by gear motors with this controller ensures sufficient operation duration for the robot.

**Table 4: Conditions in each iteration simulation.**

| Iteration # | Desired robot velocity, $V_d$(m/s) | Initial value for $\phi$(degree) | Initial value for $\psi$(degree) |
|---|---|---|---|
| 1 | 0.1 | -10 | +10 |
| 2 | 0.3 | +20 | +20 |
| 3 | 0.5 | -23 | -25 |

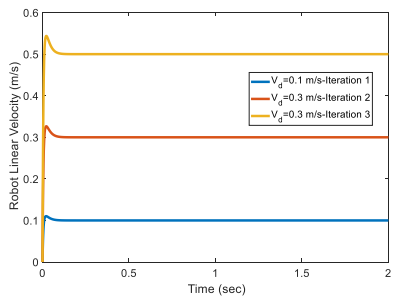

(a)

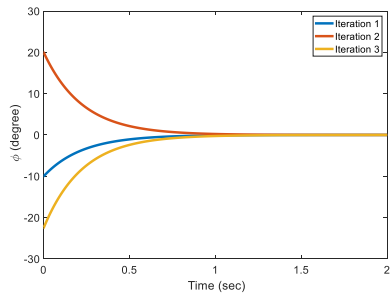

(b)

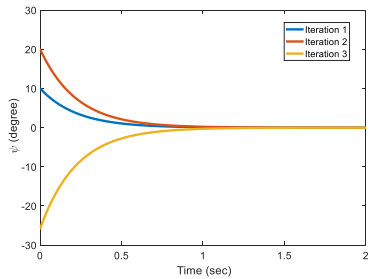

(c)

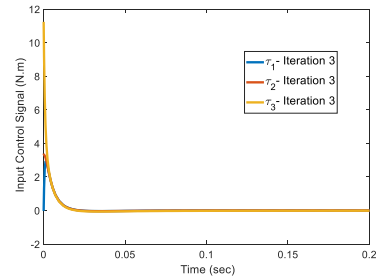

(d)

**Figure 4:** Performance of the stabilizer and velocity tracking controller in three iterations. a) Linear velocity of the robot. b) Performance of the controller in stabilizing $\boldsymbol{\phi}$ with three different initial values. c) Performance of the controller in stabilizing $\boldsymbol{\psi}$ with three different initial values. d) Input control signals, $[\boldsymbol{\tau_1} \quad \boldsymbol{\tau_2} \quad \boldsymbol{\tau_3}]^T$, in iteration 3.

## 6    CONCLUSIONS

In-pipe robots for inspection in large pipes of water distribution systems lack motion efficiency in stabilizing themselves in the pressurized environment of pipelines while tracking the desired velocity. In this paper, we model our previous under-actuated in-pipe robot and propose a stabilizer and velocity tracker controller based on linear quadratic regulator (LQR) controller and proportional integral derivative (PID) controller and validated its performance with simulation in MATLAB Simulink environment. The results show the controller can stabilize the robot inside the pipeline while the robot tracks a defined velocity.